\documentclass[10pt,twocolumn,letterpaper]{article}

\usepackage{cvpr}              

\usepackage[dvipsnames]{xcolor}

\definecolor{cvprblue}{rgb}{0.21,0.49,0.74}
\definecolor{cdbgreen}{rgb}{0.18,0.55,0.34}
\usepackage[pagebackref,breaklinks,colorlinks,citecolor=cdbgreen]{hyperref}

\usepackage{pgfplots} 

\usepackage{tikz}
\usepackage{xcolor}
\definecolor{blue}{RGB}{60,132,196}
\definecolor{red}{RGB}{207,78,56}
\definecolor{gray}{RGB}{146,146,161}
\definecolor{green4}{RGB}{46, 139, 87}
\usepackage{multirow} 
\usepackage{colortbl}
\definecolor{lightgray}{gray}{0.9}
\usepackage{hhline} 
\usepackage{pdflscape}
\usepackage{stfloats}

\definecolor{pink}{RGB}{115, 140, 166}
\definecolor{BarrierColor}{RGB}{25, 162, 30}
\definecolor{BicycleColor}{RGB}{79, 241, 199}
\definecolor{BusColor}{RGB}{198, 1, 90}
\definecolor{CarColor}{RGB}{150, 232, 149}
\definecolor{ConstVehColor}{RGB}{147, 83, 113}
\definecolor{MotorcycleColor}{RGB}{143, 43, 251}
\definecolor{PedestrianColor}{RGB}{9, 41, 148}
\definecolor{TrafficConeColor}{RGB}{86, 155, 103}
\definecolor{TrailerColor}{RGB}{118, 122, 211}
\definecolor{TruckColor}{RGB}{46, 6, 61}
\definecolor{DrivableSurfColor}{RGB}{185, 217, 16}
\definecolor{OtherFlatColor}{RGB}{142, 44, 241}
\definecolor{SidewalkColor}{RGB}{242, 218, 92}
\definecolor{TerrainColor}{RGB}{225, 12, 72}
\definecolor{ManmadeColor}{RGB}{111, 209, 144}
\definecolor{VegetationColor}{RGB}{182, 34, 111}
\usepackage{ragged2e}
\usepackage{multicol}

\def\paperID{****} 
\def\confName{CVPR}
\def\confYear{2024}

\title{AdaOcc: Adaptive Forward View Transformation and Flow Modeling for 3D Occupancy and Flow Prediction}

\author{
Dubing Chen$^{1}$\thanks{This work was partly done during internship in Inceptio.},
Wencheng Han$^{1}$,
Jin Fang$^{2}$,
Jianbing Shen$^{1}$\thanks{Corresponding author: \textit{Jianbing Shen}.
This work was supported in part by the FDCT grants 0102/2023/RIA2 and 0154/2022/A3.}, \\
$^1$SKL-IOTSC, CIS, University of Macau.\\
$^2$Inceptio Tech.\\
 {\tt\small \{dobbin.chen, wencheng256, fangjin19900820\}@gmail.com,  jianbingshen@um.edu.mo} \\
}

\begin{document}
\maketitle
\begin{abstract}
In this technical report, we present our solution for the Vision-Centric 3D Occupancy and Flow Prediction track in the nuScenes Open-Occ Dataset Challenge at CVPR 2024. Our innovative approach involves a dual-stage framework that enhances 3D occupancy and flow predictions by incorporating adaptive forward view transformation and flow modeling. Initially, we independently train the occupancy model, followed by flow prediction using sequential frame integration. Our method combines regression with classification to address scale variations in different scenes, and leverages predicted flow to warp current voxel features to future frames, guided by future frame ground truth. Experimental results on the nuScenes dataset demonstrate significant improvements in accuracy and robustness, showcasing the effectiveness of our approach in real-world scenarios. Our single model based on Swin-Base ranks second on the public leaderboard, validating the potential of our method in advancing autonomous car perception systems.
\end{abstract}    
\section{Introduction}
\label{sec:intro}

3D occupancy and flow prediction \cite{tong2023scene,wang2023openoccupancy,tian2024occ3d,liu2023fully} are critical components of autonomous driving perception systems. They involve determining the occupancy status, semantic class, and future position of each voxel in a 3D voxel space. These predictions provide rich semantic and geometric information, crucial for understanding and navigating complex driving environments. The CVPR24 occupancy and flow prediction competition \cite{tong2023scene} focuses on developing new algorithms that predict occupancy and flow solely from camera input during inference, offering a significant platform for advancing state-of-the-art 3D occupancy and flow prediction algorithms.

Our approach in this competition emphasizes innovative model design. We developed a two-stage framework to separately predict occupancy and flow. In the first stage, we train the occupancy model independently. We propose an adaptive forward view transformation method to enhance the adaptability of depth-based LSS. In the second stage, we train the flow model based on the occupancy model from the first stage. We introduced a novel sequential prediction method, using adjacent frames as inputs for the flow network. We combined regression with classification to predict flow, addressing the issue of varying flow scales in different scenes. Additionally, we use the predicted flow to warp the current frame's voxel features to the future frame and supervise using the future frame's ground truth, further enhancing prediction accuracy.

Our approach achieves an Occ Score of 0.453 without any post-hoc process, achieving 2nd place in this challenge.

\section{Our Solution}

\begin{figure*}[t]
    \centering
    \setlength{\abovecaptionskip}{0pt}
    \includegraphics[width=1\linewidth]{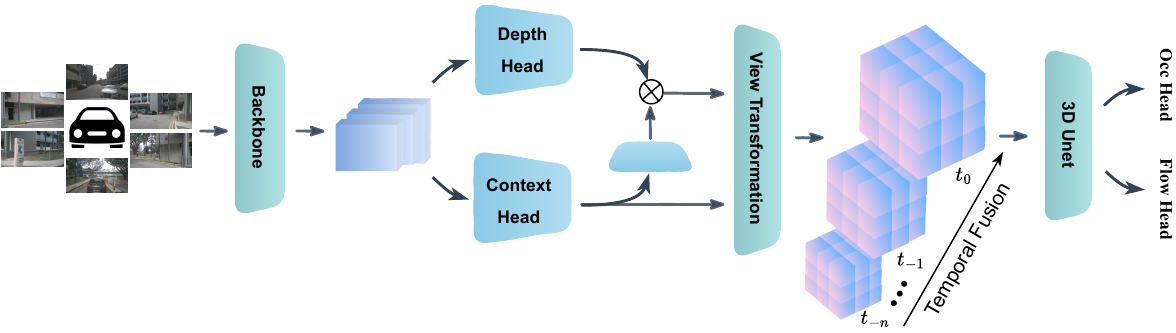}
    \vspace{0.1ex}
    \caption{An illustration of our overall pipeline, including image backbone, view transformation from 2D to 3D, Unet for 3D encoding, and task-specific heads.}
    \vspace{-1ex}
    \label{fig:main}
\end{figure*}

This section will present our solution in detail. Inspired by the general vision-based occupancy prediction pipeline, our model comprises three main components: voxel feature encoding, occupancy prediction head, and flow prediction head. We elaborate on the design of our model in Sec. \ref{sec:enc} and Sec. \ref{sec:head}, and introduce further enhancement techniques in Sec. \ref{sec:further}.

\subsection{Voxel Feature Encoding}
\label{sec:enc}
We first extract image features, then employ Lift-Splat-Shoot (LSS) \cite{philion2020lift} to transform the 2D features into 3D space. Previous methods typically use depth probability \cite{li2022bevdepth} as weights for LSS, which weakens its adaptability due to its unimodal distribution. To enhance adaptability, we integrate semantic information into the depth probability. We supervise the depth using LiDAR points and apply segmentation loss to the image features. Given the initial sparse nature of voxel features obtained via LSS, we employ the inverse process of trilinear interpolation to densify these 3D voxel features, further enhancing the adaptability of the forward view transformation. Temporal information from history frames is then fused into these voxel features, based on the sequential temporal fusion method \cite{park2022time}. We utilize a 3D Unet \cite{cciccek20163d} similar to BEVDet \cite{huang2021bevdet} for encoding 3D features. Fig.~\ref{fig:main}
is an illustration of our whole framework.

\subsection{Occupancy and Flow Prediction Heads}
\label{sec:head}
For the occupancy prediction head, we adopt a per-mask classification approach similar to Mask2Former \cite{cheng2022masked}. The mask for each category is predicted and supervised with binary cross-entropy and dice loss. We use a similar loss function for image semantic segmentation prediction.

\begin{figure}[t]
    \centering
    \setlength{\abovecaptionskip}{0pt}
    \includegraphics[width=1\linewidth]{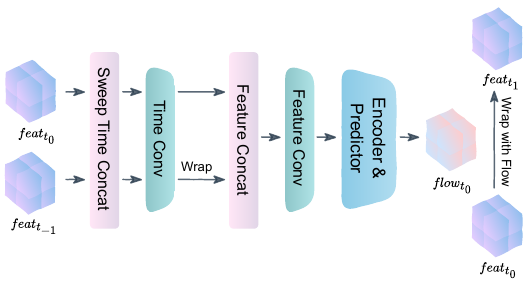}
    \vspace{0.1ex}
    \caption{Framework of the flow head. \textbf{Note} that the feature of the last frame is sequentially predicted without extra computation. The voxel feature is wrapped to the coordinates of the next frame with the predicted flow for further supervision.}
    \vspace{-1.ex}
    \label{fig:flow_head}
\end{figure}

In the flow prediction head, we predict flow using the current and previous frames as input. Rather than computing voxel features for both frames during a single inference, we sequentially utilize the previous frame's features, reducing computational overhead. Considering the significant scale variations of flow in different scenes (e.g., the flow values range from a maximum of 19.11 to a minimum of -22.73 in the OpenOcc dataset), neural networks struggle to adapt to data with such high variance. Predicting flow values for all instances becomes challenging for the network. Therefore, we transform the regression problem into a combination of classification and regression, alleviating the prediction burden on the network. We model the flow predictions within a scene into discrete adaptive bin predictions \cite{bhat2021adabins} and adaptive weights prediction. We first average the features in the scene to predict the bin centers. After defining the number of bins \( n \), we predict \( n \) probabilities using softmax and calculate each bin's center using cumulative probability:
\begin{figure}[t]
    \centering
    \setlength{\abovecaptionskip}{0pt}
    \includegraphics[width=1.05\linewidth]{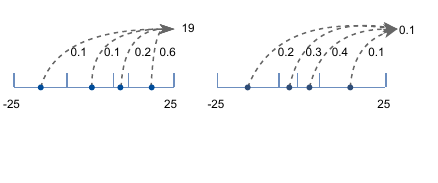}
    \vspace{-10.ex}
    \caption{An illustration of aggregating adaptive bins and adaptive weights to flows.}
    \label{fig:adabin}
    \vspace{-0.5ex}
\end{figure}
\[
c(b_i) = f_{\text{min}} + (f_{\text{max}} - f_{\text{min}})\left(\frac{b_i}{2} + \sum_{j=1}^{i-1} b_j\right) ,
\]
where \( b_i \) is the bin probability, and \( f_{\text{min}} \) and \( f_{\text{max}} \) are the pre-defined flow range. We then predict the probability \( p_k \) for each flow bin and compute the final flow prediction by weighting these bin centers:
\[
f = \sum_{k=1}^{N} c(b_k)p_k .
\]
As shown in Fig.~\ref{fig:adabin}, the prediction of different scales of flows can be transferred to different bins and weights prediction. We supervise the flow scale and direction by minimizing the L2 loss and maximizing cosine similarity with the ground truth values, respectively. Using the predicted flow, we transform the current frame's features to the next frame and supervise them with the ground truth occupancy of the future frame. To address the issue of gradient discontinuity when mapping features by coordinates, we again use the trilinear interpolation's inverse process. Supervision is applied using simple cross-entropy. Fig.~\ref{fig:flow_head} is an illustration of the flow head.

We observed that jointly predicting occupancy and flow significantly degraded occupancy performance. Thus, we adopted a two-stage training strategy: first, we trained the feature encoding module and occ head jointly with occupancy prediction, and then fine-tuned the flow head with the occupancy backbone frozen.

\subsection{Further Improvements}
\label{sec:further}
\begin{figure}[t]
    \centering
    \setlength{\abovecaptionskip}{0pt}
    \includegraphics[width=1.\linewidth,trim={0.cm 0 0 0}]{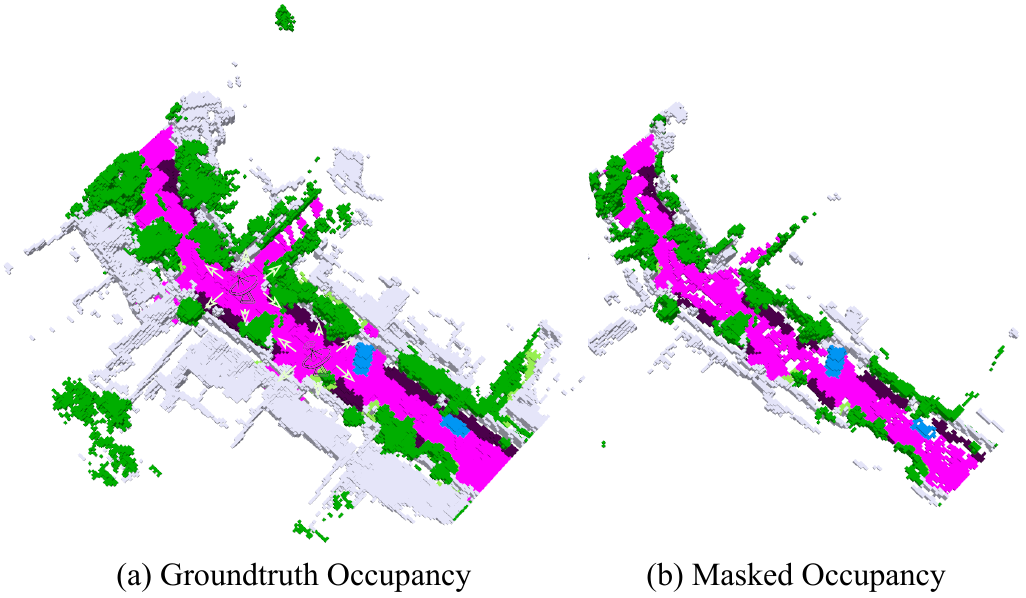}
    \vspace{0.2ex}
    \caption{Visualization of the ray visible mask. (a): Groundtruth occupancy map; (b): Traffic-critical regions of the groundtruth occupancy map.}
    \label{fig:mask}
    \vspace{-1ex}
\end{figure}

\paragraph{Ray Visible Mask.}  Beyond traffic-related factors on the road, there are numerous unnecessary elements such as buildings and vegetation far away from the ego vehicle. RayIoU \cite{liu2023fully} increases the focus on important traffic targets by setting a virtual LiDAR along the vehicle's driving path and only assessing the regions scanned by this virtual LiDAR. Therefore, we adopt this idea during training, directing more attention to the critical factors of traffic. We follow the calculation of RayIoU by setting multiple LiDAR origins along the driving path and calculating the visible regions from these origins to obtain a ray visible mask. To differentiate occupied voxels from surrounding points, we also consider voxels within 2 meters of ray-visible occupied points as critical regions. Per-frame training losses are only calculated on the critical regions with the mask.  The critical regions are visualized in Fig. \ref{fig:mask}. During training, we implement a hard example mining strategy, focusing on the training of difficult voxels based on their uncertainty. We have released mask data for public research purposes \footnote{\label{note1}\url{https://drive.google.com/file/d/10jB08Z6MLT3JxkmQfxgPVNq5Fu4lHs_h/view}.}.
\vspace{-1.6ex}
\paragraph{Stronger Setting.} We upgrade our base model from ResNet-50 \cite{resnet} to Swin Base \cite{liu2021swin}, increasing voxel feature channel sizes and using larger image input resolution.

\section{Experiment}
In this section, we discuss the dataset and metrics, provide implementation details, and present the results of ablation studies. Tab. \ref{tab:test} displays the final results on the test server.

\begin{table*}[t]
    \centering
  \resizebox{0.85\textwidth}{!}{

    \begin{tabular}{ll|cccc}
        \toprule
        & Method & IoU@1 & IoU@2 & IoU@4 & RayIoU  \\
        \midrule 
       - &Competition Baseline~\cite{li2022bevformer} & 0.196 & 0.248& 0.284& 0.243\\
        \midrule
        1 & Baseline [Bevdet, ResNet-50, 256$\times$704] & 0.268 & 0.325& 0.362& 0.318\\
        2 & Exp\#1+Mask-Based Loss & 0.296& 0.376& 0.430 & 0.367\\
        3 & Exp\#2+Segmentation Supervision, Depth Semantic Fusion & 0.310 & 0.391 & 0.444 & 0.382 \\
        4 & Exp\#3+History Fusion & 0.365& 0.444 & 0.490& 0.433\\
        5 & Exp\#3+Ray Visible Mask V1 &  0.325 & 0.399 & 0.447& 0.39\\
        6 & Exp\#3+Ray Visible Mask V2 & 0.326& 0.404& 0.453& 0.394\\
        7 & Exp\#6+History Fusion, Swin Base, 640$\times$1600, Channel 100 & 0.407& 0.483& 0.523& 0.471\\
        \bottomrule
    \end{tabular}
    }
    \caption{The occupancy prediction results \wrt different settings on the nuScenes-OpenOcc \texttt{val} set.}
    \label{tab:experiment_results}
\end{table*}

\begin{table*}[t]
\centering
\resizebox{0.92\textwidth}{!}{
\begin{tabular}{c|c|c|c|c|c}
\toprule

Experiment& OCC Score & RayIoU & mAVE@TP & mAVE@LQ &mAVE@Per-voxel \\
\midrule
Swin Base w/o Flow& - & 0.471 & -& -&-  \\
Swin Base + Flow Joint Train & 0.443  & 0.43.5 & 0.486 &0.955 &0.508\\
Swin Base (fix) + Flow & 0.479  & 0.471 & 0.467 &0.914 &0.541 \\
Swin Base (fix) + AdaFlow, Future frame Sup& 0.481  & 0.471 & 0.457 &0.908 &0.534 \\
\bottomrule
\end{tabular}
}
\caption{The flow prediction results \wrt different settings on the nuScenes-OpenOcc \texttt{val} set. mAVE@TP is the original flow metric of the challenge. We introduced two additional metrics: per-voxel mAVE (mAVE@Per-Voxel), and mAVE for all points queried from the LiDAR origin (mAVE@LQ).}
\vspace{-1.5ex}
\label{tab:flow_ab}
\end{table*}

\begin{table}[t]
\centering
\resizebox{0.478\textwidth}{!}{
\begin{tabular}{c|c|c|c|c|c}
\toprule

 OCC Score & RayIoU@1 & RayIoU@2 & RayIoU@4 & RayIoU &mAVE \\
\midrule
0.453 & 0.398 & 0.459 &0.496& 0.451& 0.529  \\

\bottomrule
\end{tabular}
}
\vspace{0.1ex}
\caption{Final results on the nuScenes-OpenOcc \texttt{test} set.}
\vspace{-3ex}
\label{tab:test}
\end{table}

\subsection{Dataset and Metrics}
The 3D Occupancy and Flow Prediction Challenge Dataset \cite{wang2023openoccupancy} is built upon the nuScenes dataset \cite{caesar2020nuscenes}. It comprises 700 sequences for training, 150 for validation, and 150 for testing. Each frame includes six surround-view images with a resolution of $900 \times 1600$. Occupancy and flow annotations are provided for each frame within the range of [-40m, -40m, -1m, 40m, 40m, 5.4m], with voxel resolution set at 0.4 meters. The dataset includes flow annotations for the x and y axes and semantic annotations for 17 categories (including ``unoccupied"). No external data is utilized in our method. The final evaluation metrics are the class-averaged RayIoU \cite{liu2023fully} across all classes and the mAVE for foreground classes. The overall evaluation score is computed as 
\[
\textrm{Occ Score}=0.9 \cdot \textrm{RayIoU} + 0.1 \cdot \max(1 - \textrm{mAVE},0).
\]

\subsection{Implementation Details}
\paragraph{Training Strategies.}
We conducted preliminary experiments using BEVDet as the baseline model, training on 8 NVIDIA A100 GPUs. We used the AdamW optimizer \cite{loshchilov2017decoupled} with a learning rate of 2e-4, a weight decay of 0.05, and a total batch size of 32. We adopted exponential moving average (EMA) \cite{huang2021bevdet} for updating model weights. When using ResNet-50 \cite{resnet} as the backbone, we trained the baseline model without temporal fusion for 24 epochs. For the baseline with temporal fusion, we followed the methodologies of SOLOFusion \cite{park2022time} and FB-BEV to apply CBGS \cite{park2022time} for 12 epochs. For the Swin Base \cite{liu2021swin} model, we trained the occ head with a total batch size of 8 and trained with CBGS for 5 epochs. The flow head was then fine-tuned for an additional 5 epochs.

\paragraph{Network Details.}
We used common data augmentation strategies \cite{huang2021bevdet}, including random flip, scale, and rotation for image and flip augmentation for voxel features. We initialized our networks using publicly available models. For ResNet-50, we used a BEVDet \cite{huang2022bevdet4d} detection pre-trained model, and for the Swin Base, we initialized with GeoMIM \cite{liu2023geomim} pre-trained on Occ3d \cite{tian2024occ3d}. The voxel size in training is set to $200\times200\times16$. When using ResNet-50 as the backbone, the input image size was $256\times704$, with voxel feature channels set to 32. The image size is set to $640\times1600$ with 100 voxel channels for the Swin Base model. We used 16 history frames for temporal fusion, following a sequentially based fusion pipeline \cite{park2022time}.

\subsection{Ablation Study for Occ and Flow Training}

We conduct preliminary experiments using the ResNet-50 backbone to quickly validate the efficacy of the proposed components, then scale up using the validated effective methods. We found that training occ and flow together decreased RayIoU by about two points. Thus, we adopted a two-stage training strategy: first train occ, then train flow. Tab. \ref{tab:experiment_results} presents the ablation study results on occ training, we incrementally added modules to the BEVDet baseline, each significantly improving performance. In baseline 2, we adopted the mask-based loss and hard example mining strategy following Mask2Former \cite{cheng2022masked}, achieving significant improvements. Similarly, in baseline 5, we used the ray visible mask to focus the network on important traffic areas, improving RayIoU. In baseline 6, performance (particularly IoU@4) is further enhanced by including voxels within 2 meters of the ray termination point in training. Additionally, in baseline 3, image segmentation loss and the injection of semantic information into depth alleviated the unimodal distribution, enhancing adaptability. Finally, our method surpassed the official baseline method with a RayIoU of 0.151 under the same image output size and backbone settings. In baselines 4 and 7, we leveraged historical frame information and larger models, which naturally led to improvements.

Tab.~\ref{tab:flow_ab} presents the evaluation results of flow head training. When fine-tuning both occ and flow on a trained occ network, we observed a significant drop in occ prediction performance. Therefore, we froze the occ network and the encoder, fine-tuning only the flow head. This preserved the occ performance while effectively predicting flow. To provide a more comprehensive comparison of flow performance, we introduced two additional metrics: one calculates a per-voxel mAVE (mAVE@Per-Voxel), and the other calculates mAVE for all points queried from the LiDAR origin (mAVE@LQ). We found that fine-tuning all parameters resulted in a decrease in the per-voxel mAVE, but did not provide gains for the other two metrics. Additionally, our proposed AdaBin method and the use of future frames for supervision improved flow prediction.

\section{Final Results and Conclusion}
In this report, we describe our solution in detail for the CVPR 2024 Autonomous Grand Challenge Track On Occupancy and Flow. The model of our final submission employs Swin Base as the image backbone. The image resolution is set to $640 \times1600$. The encoded voxel size is $200 \times 200 \times 16$, with a channel size of 100. We use 16 historical frames. No post-processing techniques, such as test-time augmentation or model ensembling, are applied. This model achieves an Occ Score of 0.453 on the test server, ranking 2nd on the test server.

{
    \small
    \bibliographystyle{ieeenat_fullname}
    \bibliography{main}
}
\onecolumn
\setcounter{page}{1}
\setcounter{figure}{0}
\section*{Apendix}

\appendix
\renewcommand\thefigure{A.\arabic{figure}} 
\renewcommand\theequation{A.\arabic{equation}} 
\renewcommand\thetable{A.\arabic{table}}

\begin{figure*}[hb]
    \centering
    \setlength{\abovecaptionskip}{0pt}
    \includegraphics[width=0.9\linewidth]{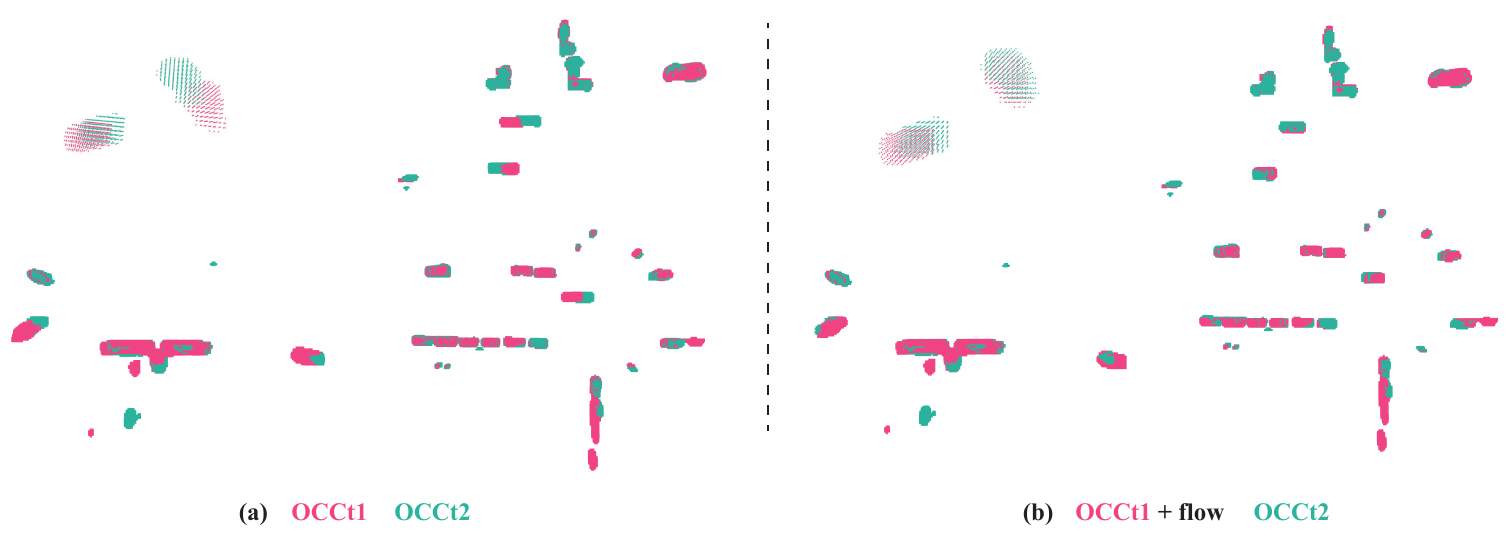}
    \vspace{1.ex}
    \caption{An illustration of our flow prediction.}
    \label{fig:flow_pred}
\end{figure*}

\begin{figure*}[hb]
\centering
    \includegraphics[width=0.95\textwidth]{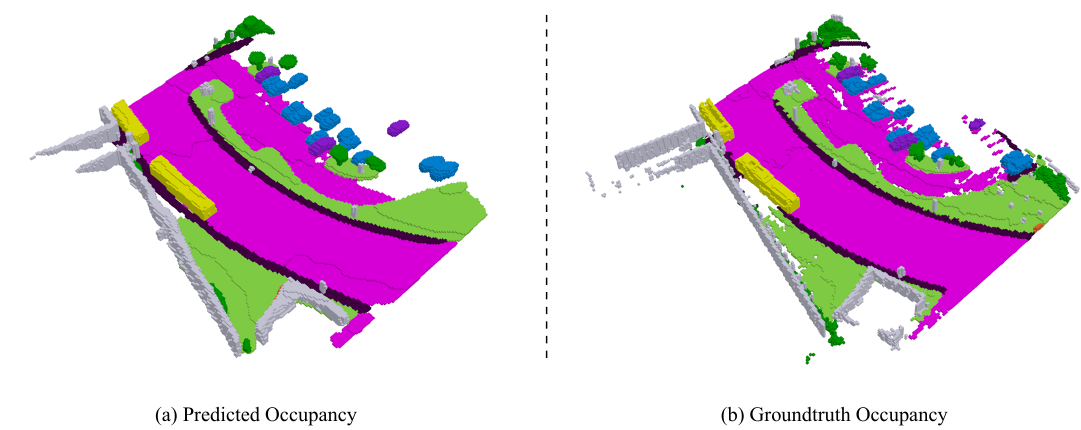}

\caption{A comprision of the predicted (left) and groundtruth occupancy map (right).}

\label{fig:occ}
\end{figure*}

\begin{multicols}{2}
As shown in Fig.~\ref{fig:flow_pred} and Fig.~\ref{fig:occ}, we visualized the comparison between the predicted and ground truth occupancy, as well as the predicted flow and its application in transforming the current frame to the next frame. The visualizations demonstrate the accuracy of our model in predicting occupancy status and semantic class. The alignment between predicted and actual occupancies confirms the effectiveness of our dual-stage framework and the integration of AdaBin. For flow prediction, the visual transformation shows our model's capability to capture temporal dynamics and spatial continuity, validating the robustness of our approach.
\end{multicols}

\end{document}